\pdfoutput=1

\documentclass[11pt]{article}

\usepackage[final]{acl}

\usepackage{times}
\usepackage{latexsym}

\usepackage[T1]{fontenc}

\usepackage[utf8]{inputenc}

\usepackage{microtype}

\usepackage{inconsolata}

\usepackage{graphicx}
\usepackage{multirow}
\usepackage{booktabs}
\title{Leveraging Context for Multimodal Fallacy Classification in Political Debates}

\author{
  Alessio Pittiglio \\
  DISI, University of Bologna, Bologna, Italy \\
  \texttt{alessio.pittiglio@studio.unibo.it}}

\begin{document}
\maketitle
\begin{abstract}
In this paper, we present our submission to the MM-ArgFallacy2025 shared task, which aims to advance research in multimodal argument mining, focusing on logical fallacies in political debates. Our approach uses pretrained Transformer-based models and proposes several ways to leverage context. In the fallacy classification subtask, our models achieved macro F1-scores of 0.4444 (text), 0.3559 (audio), and 0.4403 (multimodal). Our multimodal model showed performance comparable to the text-only model, suggesting potential for improvements.

\end{abstract}

\section{Introduction}

Politicians have always resorted to stratagems in an attempt to convince as many people as possible to vote for them. In recent years, there have been initiatives aimed at verifying the truthfulness of politicians' statements. However, this type of verification, although useful, is not sufficient: many persuasive techniques do not rely on false facts, but on misleading reasoning, such as logical fallacies.

We address the MM-ArgFallacy2025 shared task\footnote{\url{https://nlp-unibo.github.io/mm-argfallacy/2025/}}, which focuses on multimodal detection and classification of argumentative fallacies in political debates. It proposes two subtasks: Argumentative Fallacy Detection (AFD) and Argumentative Fallacy Classification (AFC). In this work, we focus on the AFC task. The classes for the AFC are those proposed by \citet{ijcai2022p575}. For both sub-tasks, three input settings are provided: text-only, audio-only, and a combination of both (text+audio). The dataset used is MM-USED-fallacy \cite{mancini-etal-2024-multimodal}, which is available via MAMKit \cite{mancini-etal-2024-mamkit}. Initially, only the training set was released, so we created our own validation split for internal evaluation. The official test set, released later without labels, was used for the final submission. The primary assessment criterion for AFC is the macro F1-score.

We propose a system that leverages Transformer-based models for both text and audio modalities. For text input, we experiment with a range of architectures, including a simple concatenation of text input and its previous context, a context pooling model, and a cross-attention model with a gating mechanism for context integration. Among these, the context pooling model combined with RoBERTa-large \citep{liu2019robertarobustlyoptimizedbert} yielded the best performance. For audio input, we use a fine-tuned HuBERT Base model \citep{hsu2021hubertselfsupervisedspeechrepresentation}, applying temporal average pooling to obtain global embeddings. We also experiment with a variant that combines embeddings from the audio input and its context, following a strategy similar to the context-pooling approach used for text. For multimodal integration, we explore ensemble strategies that combine the outputs of the text and audio models using weighted averaging and majority voting. The main contribution of this work is an investigation into how context from previous sentences can be effectively incorporated across both modalities. While textual context consistently improved performance, the impact of audio context was less significant. Finally, we suggest some possible directions for future improvements. Our code is publicly available \footnote{\url{https://github.com/alessiopittiglio/mm-argfallacy}}.

\section{Related Work}

Until a few years ago, most research in the field of argument mining in political debates focused exclusively on text data. More recently, it has been shown that the use of audio data can also be informative. In \citet{Lippi_Torroni_2016}, the authors introduced a dataset based on the 2015 UK elections and demonstrated that the inclusion of audio features improves claim extraction.

In \citet{haddadan-etal-2019-yes}, the authors addressed the problem of recognizing argumentative components in political debates. They created a new corpus, USElecDeb60To16, and highlighted that a key factor in argumentative mining is the use of context.

In \citet{ijcai2022p575}, the dataset proposed by \citet{haddadan-etal-2019-yes} was extended with annotations for six fallacy categories: Ad Hominem, Appeal to Authority, Appeal to Emotion, False Cause, Slogan, and Slippery Slope. A model based on Longformer \cite{DBLP:journals/corr/abs-2004-05150} was proposed. The model was trained to classify fallacies using debate context, fallacy snippets, and argumentative components and relations from the original dataset.

Recent studies in multimodal argument mining have highlighted the benefits of combining text and audio inputs for improved performance. In \citet{mancini-etal-2022-multimodal}, the authors showed that features extracted from pretrained models outperform traditional features (e.g., MFCCs) in tasks such as claim detection and argumentative relation classification. However, their effective integration with text inputs remains an open problem. \citet{mancini-etal-2024-multimodal} introduced MM-USED-fallacy, the first multimodal corpus for classifying argumentative fallacies, extending the USED-fallacy dataset \citep{ijcai2022p575}. Their proposed architecture combines pretrained models for text and audio using a fusion approach, achieving significant improvement over text-only models like BERT \citep{DBLP:journals/corr/abs-1810-04805} and RoBERTa \citep{liu2019robertarobustlyoptimizedbert}.

Despite these advances, the usage of context in multimodal argument mining remains unexplored. Prior work has primarily focused on classifying sentences without evaluating how context from preceding sentences influences performance across modalities. In this work, we address this gap by investigating context-aware architectures for both text and audio, as part of the AFC shared task.

\section{Input Settings}

For the challenge, we had access only to the training set. The train split we worked on was composed of 1278 sentences. It is relatively small and highly imbalanced dataset. To prepare a validation set, we performed an 80/20 train/validation split at the sentence level, using the stratify option to keep the same class distribution as the original dataset.

\subsection{Text}

\paragraph{Data processing.}

The text was not preprocessed in any way. We directly used the tokenizer of each backbone model to tokenize the text.

\paragraph{Data encoding.}

We used a range of backbones varying in size. These include smaller models such as BERT-base \cite{DBLP:journals/corr/abs-1810-04805}, RoBERTa-base \citep{liu2019robertarobustlyoptimizedbert}, and DeBERTaV3-base \citep{he2023debertav3improvingdebertausing}, as well as larger ones like RoBERTa-large, DeBERTaV3-large, and the recent ModernBERT-large \citep{warner2024smarterbetterfasterlonger}. The latter integrates new features, such as FlashAttention 2 \citep{dao2023flashattention2fasterattentionbetter}, and has the largest context window (4080 tokens). Compared to other backbones like Longformer \cite{DBLP:journals/corr/abs-2004-05150}, ModernBERT-large was chosen for its more efficient training. All encoders were fine-tuned with all layers unfrozen.

\paragraph{Architectures.}

To incorporate context, three approaches were used:

\begin{enumerate}
    \item \textbf{Concat}, involving the concatenation of the text and its context, separated by the tokenizer's separator token.
    \item \textbf{ContextPool}, based on pooled embeddings obtained from a shared transformer to classify the text based on its context.
    \item \textbf{CrossAttn}, which uses a cross-attention mechanism to integrate text and context, followed by a gating mechanism for fusion.
\end{enumerate}

See Appendix~\ref{subsec:architectures-details} for more details about the architectures. The model that performed better was \textbf{ContextPool-RoBERTa}, which combines the ContextPool architecture with RoBERTa-large. To maximize performance, we submitted an ensemble composed of our three best models trained using this successful combination. Two of the models were trained with the same hyperparameters, while the third used a larger context window size of 5, with all other parameters kept identical (see Table~\ref{tab:hyperparams_text}). At inference time, predictions from the three models were combined by averaging their predicted logits, using the weights reported in Table~\ref{tab:hyperparameters_ensemble_text}, which were obtained via Bayesian optimization.

\subsection{Audio}

\paragraph{Data processing.}

For the audio processing, we used a custom processor component, based on the one from the MAMKit library \cite{mancini-etal-2024-mamkit}, to handle the audio and its context. We implemented a function that returns 100 ms of silence for empty inputs. Valid audio files are loaded, resampled to 16 kHz, and concatenated along the time axis if multiple files are provided. Instead of using the built-in collator provided by MAMKit, which was implemented for training another transformer starting from the features extracted from a backbone, we decided to fine-tune a model directly for this task. Therefore, we implemented a new collator specifically for this purpose. Each input is optionally truncated to a maximum length of 15 seconds. This is done because processing very long audio causes an out-of-memory error. When context is available, it is processed in the same way.

\paragraph{Data encoding.}

We evaluated the following backbones, also used by \citet{mancini-etal-2024-multimodal}: Wav2Vec 2.0 Base \cite{DBLP:journals/corr/abs-2006-11477} fine-tuned on 960 hours of LibriSpeech, WavLM Base+ \cite{DBLP:journals/corr/abs-2110-13900} fine-tuned on 100h of LibriSpeech clean, and HuBERT Base \cite{hsu2021hubertselfsupervisedspeechrepresentation}. Despite similar training setups, Wav2Vec 2.0 Base and WavLM Base+ did not perform well in early experiments, while HuBERT Base showed promising results.

\paragraph{Architectures.}

We used two architectures. The first architecture is \textbf{HuBERT-Base fine-tuned}, which is simply a fine-tuned version of the backbone. Our idea was to unfreeze only some layers. With audio, we cannot use the same mechanism used for text (pooling embeddings), so we used the temporal average (average along the sequence dimension) of the embeddings. The result is a global embedding for each audio sample without completely losing the temporal dimension. Inspired by the results obtained with text, the second architecture we implemented is \textbf{TemporalAvg} which combines the mean pooling of the audio snippet with that obtained from the audio of the context.

\subsection{Text-Audio}

In order to obtain the best possible results, we decided to create an ensemble of our best text model (ContextPool-RoBERTa) and audio model (HuBERT-Base fine-tuned). Taking an arithmetic average of the logits of each model is the simplest way to do it, but there are more effective methods to improve it. One such method is to use a weighted average. The optimization was performed using the Bayesian optimization technique \cite{snoek2012practicalbayesianoptimizationmachine}. The metric we aimed to maximize was the F1 score calculated on the validation set. The validation set used was the same one used during the training of the models. Each model was trained using the same train/validation split; otherwise, this could distort the metric. The optimization was performed with 20 iterations and 15 initial points (see Table \ref{tab:hyperparameters_ensemble_text_audio} for the final weights). Additionally, we tested a technique based on majority voting across three models. These included: (1) our best text-only model (ContextPool-RoBERTa) (2) our best audio-only model (HuBERT-Base fine-tuned), and (3) the ensemble combining our best text and audio models. This majority voting ensemble was used for the final submission.

\section{Experimental Setup}

\subsection{Context Usage}

Following the definition provided in the challenge specifications, we define the \textit{debate context} of a given input as the sequence of previous sentences in the debate, aligned with their corresponding audio segments. For a sentence at index $i$ in a political debate, the context consists of all preceding sentences, i.e., those from 0 to $i - 1$, where 0 denotes the first sentence in the debate. For text, we experimented with the three previously mentioned architectures: Concat, ContextPool and CrossAttn. For the audio, we used only TemporalAvg, as it is an adaptation of ContextPool, the one that performs best on text. To adapt ContextPool for audio, we slightly modified the architecture, while maintaining the core idea of concatenating audio and context. Since Audio Transformers downsample raw waveforms into shorter sequences, masking padded tokens isn't directly possible; instead, we apply average pooling over time. To assess the contribution of context, we conducted ablation studies across different configurations (see Table~\ref{tab:ablation_text} and Table~\ref{tab:ablation_audio}). We tested all combinations of window sizes and architectures, with window sizes ranging from 1 to 6.

\begin{table*}[]
\centering
\begin{tabular}{llllllll}
\toprule
\textbf{Architecture}     & \textbf{N=0 (No Ctx)} & \textbf{N=1}    & \textbf{N=2}    & \textbf{N=3}    & \textbf{N=4}    & \textbf{N=5}    & \textbf{N=6}    \\ \hline
Transformer (No Context)  & 0.6131                & -               & -               & -               & -               & -               & -               \\
Concat                    & -                     & 0.5538          & 0.5932          & 0.4767          & 0.5331          & 0.5941          & 0.5431          \\
ContextPool               & -                     & \textbf{0.6636} & 0.6479          & 0.5786          & \textbf{0.6983} & 0.6542          & \textbf{0.6494} \\
CrossAttn                 & -                     & 0.5699          & 0.6086          & 0.5395          & 0.6304          & 0.6344          & 0.6219          \\
w/ Gate                   & -                     & 0.5867          & 0.6171          & 0.5487          & 0.5388          & 0.6181          & 0.6032          \\
w/ Attentive Pooling      & -                     & 0.6603          & \textbf{0.6527} & 0.6227          & 0.6590          & \textbf{0.6579} & 0.6383          \\
w/ Gate \& Attentive Pool & -                     & 0.6280          & 0.6520          & \textbf{0.6407} & 0.6261          & 0.3701          & 0.6377          \\
\bottomrule
\end{tabular}
\caption{Ablation study on context integration strategies and window size (N) for text modality (F1-Macro).}
\label{tab:ablation_text}
\end{table*}

We observed that the Concat approach did not yield improvements over the model without context. ContextPool achieved the highest F1 score with a context size of $N = 4$. One interesting observation is that the improvement does not scale linearly with increasing context. For instance, ContextPool performance slightly drops after $N = 4$. 

Vanilla CrossAttn remains below the baseline until $N = 3$. After that, adding context becomes beneficial. With CrossAttn using attentive pooling, we observe immediate improvements that remains relatively stable across different values of $N$. It demonstrates performance comparable to ContextPool but does not reach its peak. In contrast, CrossAttn with gate fusion and attentive pooling shows more inconsistent behavior, achieving the lowest overall score at $N = 5$. 

For the audio modality, the ablation results reported in Table~\ref{tab:ablation_audio} reveal that adding context does not consistently improve performance. The TemporalAvg architecture shows fluctuating F1 scores across different values of $N$, with no clear rising trend. The baseline Hubert model, fine-tuned without additional context, outperforms all TemporalAvg configurations.

\begin{table*}[]
\centering
\begin{tabular}{llllllll}
\toprule
\textbf{Architecture} & \textbf{N=0 (No Ctx)} & \textbf{N=1} & \textbf{N=2} & \textbf{N=3} & \textbf{N=4} & \textbf{N=5} & \textbf{N=6} \\ \hline
Hubert Base fine-tuned             &  0.48061               & -            & -            & -            & -            & -            & -            \\
TemporalAvg           & -                     & 0.4282       & 0.4149       & 0.3856       & 0.4460       & 0.4553       & 0.4518       \\
\bottomrule
\end{tabular}
\caption{Ablation study on context window size (N) for audio modality (F1-Macro).}
\label{tab:ablation_audio}
\end{table*}

\subsection{Training}

The training was conducted using an NVIDIA RTX 3090 GPU. When available, we utilized FlashAttention \cite{dao2023flashattention2fasterattentionbetter} to accelerate training. To investigate whether a positive correlation exists between the input and the context window size, we experimented with varying context lengths. The models were trained with bf16 mixed precision, and AdamW \cite{loshchilov2019decoupledweightdecayregularization} was used as the optimizer. A linear learning rate scheduler was used, with the warmup phase set to 30\% of the total training steps. To mitigate overfitting and prevent wasting computational resources, early stopping based on the validation loss with a patience of 5 was applied during all training runs. For class weights, we used those also used by \citet{mancini-etal-2024-multimodal}. We also attempted to dynamically compute class weights from our training split but observed no significant improvements. For each model, we tested a range of learning rates, determined based on the model's response to an initial learning rate. Each learning rate was evaluated across three independent experiments with different random seeds, and the optimal rate was selected based on the average performance on the validation set. Hyperparameter tuning was conducted using W\&B Sweeps \cite{wandb}. A detailed list of hyperparameters and training configurations is provided in Appendix~\ref{subsec:training_details} (see Table~\ref{tab:hyperparams_text} and Table~\ref{tab:hyperparams_audio}).

\section{Results}

The results for the final submitted models are reported in Table~\ref{tab:external_eval}. Observing the ranks, our audio model (HuBERT-Base fine-tuned) performs particularly well compared to other models. Moreover, the multimodal model did not outperform the text and audio models. We hypothesize that the lack of interaction between modalities during training may have limited the model's ability to exploit cross-modal correlations, thereby reducing potential gains. Overall, We conclude that our technique for combining text and audio was not effective.

\begin{table}[]
  \centering
  \small
  \begin{tabular}{lll}
    \toprule
    \textbf{Input}              & \textbf{Team}            & \textbf{F1}       \\ \hline
    \multirow{3}{*}{Text-Only}  & Team NUST                & 0.4856            \\
                                & Baseline BiLSTM          & 0.4721            \\
                                & Our team                 & 0.4444            \\ \hline
    \multirow{3}{*}{Audio-Only} & Our team                 & 0.3559            \\
                                & Team EvaAdriana          & 0.1858            \\
                                & Team NUST                & 0.1588            \\ \hline
    \multirow{3}{*}{Text-Audio} & Team NUST                & 0.4611            \\
                                & Our team                 & 0.4403            \\
                                & Baseline RoBERTa + WavLM & 0.3816            \\ 
  \bottomrule
  \end{tabular}
  \caption{External evaluation of submissions on the test set. Reported F1 scores are macro-averaged.}
  \label{tab:external_eval}
\end{table}

Since no detailed results for each class were provided, we evaluated the same model checkpoints on our validation set to analyze class performance (Table~\ref{tab:eval_class_specific}). \textit{Appeal to Emotion} achieves the highest F1 score for both text and audio, likely due to its high concentration in the dataset. In the text modality, the model also performs well on \textit{Slippery Slope}, likely benefiting from lexical patterns that indicate causality or escalation. However, performance drops on \textit{False Cause}, probably due to the complexity of the reasoning required.

In the audio modality, the second highest class is \textit{Slogan}, as it is easily recognizable and strongly dependent on vocal pitch. In contrast, the model struggles with \textit{Slippery Slope}, where increased use of context might improve performance. However, audio was truncated to 15 seconds due to memory constraints.

Interestingly, the text-audio model performs like a "faded" version of the text-only model. This suggests that the current fusion approach may dilute strong unimodal signals rather than enrich them with additional information. Improvement is observed only in \textit{Slogan}, indicating that integration can be beneficial when modalities contribute complementary.

This provides fundamental insight into the fact that features are orthogonal across modalities, suggesting that a more complex fusion strategy might better leverage the strengths of each.

\begin{table}[]
  \centering
  \begin{tabular}{llll}
    \toprule
    \textbf{Class} & \textbf{Text-only} & \textbf{Audio only} & \textbf{Text-audio} \\ \hline
    \textbf{AE}    & 0.8802             & 0.7616              & 0.8519              \\
    \textbf{AA}    & 0.7105             & 0.3636              & 0.6667              \\
    \textbf{AH}    & 0.6909             & 0.4444              & 0.6667              \\
    \textbf{FC}    & 0.6316             & 0.3636              & 0.6316              \\
    \textbf{SS}    & 0.7500             & 0.3333              & 0.7500              \\
    \textbf{S}     & 0.6667             & 0.5455              & 0.7143              \\ 
  \bottomrule
  \end{tabular}
  \caption{F1 scores for each class on the validation set. AE: Appeal to Emotion, AA: Appeal to Authority, AH: Ad Hominem, FC: False Cause, SS: Slippery Slope, S: Slogans}
  \label{tab:eval_class_specific}
\end{table}

\section{Conclusion}

Our main contribution is having explored how to leverage information from previous sentences. Our second approach, ContextPool-RoBERTa, proved more effective than the other methods we tested. In contrast, for the audio setting, we successfully trained a model (HuBERT-Base fine-tuned) capable of distinguishing different fallacies. Furthermore, our decision to fine-tune a model and unfreeze certain layers proved more efficient and merits further exploration. We believe the reason for this efficiency lies in the improvement of the feature extractor when fine-tuned on the specific argument domain.

We also explored a late fusion approach to combine the predictions and majority voting in an attempt to improve performance. However, we found this technique to be less effective. Additionally, we noted that features learned from the text and audio models capture distinct aspects; thus, further exploration of techniques to combine these features in a more complex manner could be promising.

\section*{Limitations}

\paragraph{Dataset.}

Duplicate samples were present in the dataset. Specifically, we identified 66 duplicate phrases (10 repeated three times, 1 repeated four times) and 16 inconsistent samples. By "inconsistent", we mean that the phrase and the context are the same, but the labels differ. While we acknowledged their presence, we did not actively resolve or mitigate their impact during training. The dataset was subsequently updated by the organizers to remove duplicates, but the 16 inconsistencies persisted.

\paragraph{Audio processing.}

Due to out-of-memory errors encountered during the training of the audio model, we adopted a truncation strategy. All audio samples in our dataset were truncated to a maximum of 15 seconds. Truncation is applied at the sample level: if the audio exceeds 15 seconds in length (corresponding to 240000 samples at a 16 kHz sampling rate), it is truncated. The same applies to the context.

This choice, although necessary for experimental feasibility, could have introduced potential implications for model performance. We conducted an analysis of the audio length distribution in the MM-USED-fallacy dataset (see Appendix~\ref{subsec:audio_length_distribution}). Although the average duration of the input samples is 9.51 seconds, 17\% of the samples exceeded the threshold, suggesting that a significant portion of the dataset was truncated. Moreover, truncating audio to 15 seconds imposed limitations on context exploitation. Table~\ref{tab:avg_duration_context} reports the average duration of the context audio as a function of the window size. For instance, a context window of 6 spans 31.44 seconds, necessitating truncation to fit the limit. This may have impaired the model's performance, particularly for fallacies like Slippery Slope, where extended context could improve classification.

Table~\ref{tab:length_distribution_classes} shows the length distribution for each class. Classes such as FC and SS have a longer average length compared to other classes. Truncating all audio inputs to 15 seconds disproportionately affects these classes, potentially discarding informative content and introducing a bias toward shorter utterances. 

One minor adjustment that could help mitigate the issue is to truncate from the beginning of the audio, as the truncation was applied to the end of the audio sequence. An empirical analysis comparing performance across different strategies for handling audio length could represent an important direction for future work. Such a study could precisely quantify the trade-off between computational efficiency and information fidelity.

\paragraph{Fusion strategy.}

The late fusion of text and audio models did not outperform individual text- or audio-only models. This suggests that the current fusion approach is suboptimal, and more advanced techniques, should be explored to better integrate modalities.

\bibliography{custom}

\appendix

\section{Appendix}
\label{sec:appendix}

\subsection{Architectures details}
\label{subsec:architectures-details}

\paragraph{Concat.}

\begin{figure}[t]
  \includegraphics[width=\columnwidth]{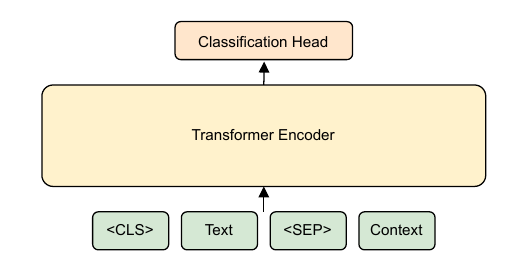}
  \caption{The Concat architecture.}
  \label{fig:concat}
\end{figure}

We use the separator token to divide the text and its context and let the model decide how to attend to each token (see Figure~\ref{fig:concat}). Whether the context appears before or after the text is a matter of choice. The downside of this approach is that it creates very long token sequences. With too much context, the transformer may lose focus on what the task requires, i.e., classifying the text. On the positive side, however, it allows the transformer to decide how to allocate attention to different tokens.

\paragraph{ContextPool.}

\begin{figure}[t]
  \includegraphics[width=\columnwidth]{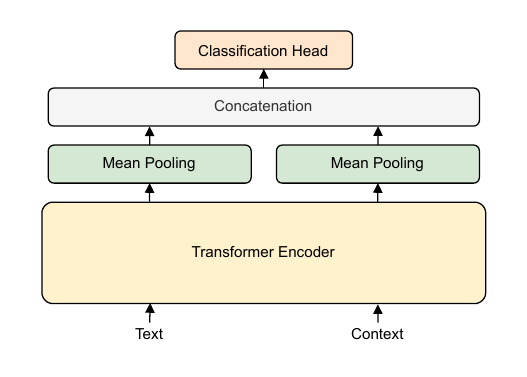}
  \caption{The ContextPool architecture.}
  \label{fig:context_pool}
\end{figure}

The idea is similar to the one presented in \citet{reimers-gurevych-2019-sentence}: a Siamese BERT-base network. Text and context pass through the encoder. Then, for both, we perform mean pooling and concatenate the information. Finally, there is a classification head (see Figure~\ref{fig:context_pool}).

\paragraph{ContextAtt.}

\begin{figure}[t]
  \includegraphics[width=\columnwidth]{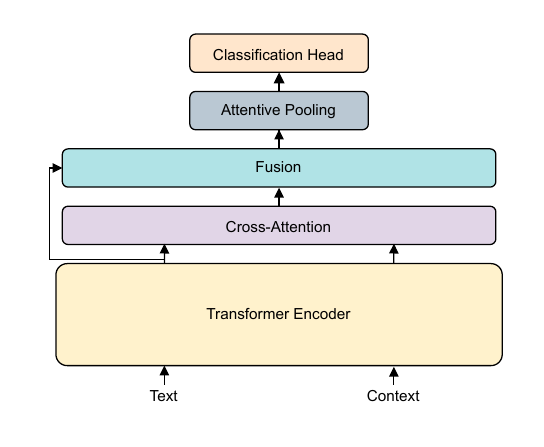}
  \caption{The ContextAtt architecture.}
  \label{fig:context_att}
\end{figure}

It's a transformer with a cross-attention mechanism for integrating text and context (see Figure~\ref{fig:context_att}). This model processes text and context inputs separately through a shared transformer, then applies cross-attention to integrate context information into the text representation. This enriched embeddings are fused with the original text embeddings through a gate and compressed into a global embedding via attentive pooling.

The fusion gate is nothing more than an MLP that takes as input the text vector and the context vector concatenated and projected into a space of dimension equal to the hidden size. A sigmoid function was then applied, returning values between 0 and 1. In this way, we obtain a mask that weights the importance of the context inputs during fusion. In our implementation, the text embedding was always assigned full weight, while the context-aware embedding provides an additive contribution modulated by the gate. Finally, a normalization layer was applied at the end of the fusion phase.

Moreover, instead of using average pooling, in this case we implemented attentive pooling, a type of pooling that allows the model to learn which tokens to attend to. For this, we created a small MLP to compute attention scores for each token. The scores are then normalized using softmax. Applying these weights to the embedding tokens, we obtained a global weighted embedding.

\subsection{Training details}
\label{subsec:training_details}

We tracked all the experiments using Weights \& Biases. To manage all the different configuration, we used YAML files. The default class \texttt{MAMKitLightingModel} has been expanded by adding more features. First, we added the support for a learning rate scheduler. Another addition was a hook to log total gradient norm at each step. This addition has been very useful in helping us understand how large the gradient norm was. Specifically, using this graph, we were able to detect that applying gradient clipping was harmful in our case, as every time we tried it, we observed very large spikes  in the gradient that damaged the learning process. For the audio, we implemented a different version of the \texttt{MAMKitLightingModel} that supports differential learning rates. This allowed us to train the head and the backbone of the model with different learning rates, even though the final model was trained using the same learning rate for both the head and the backbone. Two schedulers were tried: a linear scheduler and a cosine scheduler, both with warmup. Also, we tried different warmups. Initially, we set it to 10\% of the total number of steps, but then, we noticed that increasing the warmup to 30\% allowed us to obtain better results. Furthermore, during model training we applied high internal precision to all float32 matrix multiplications, trading off precision for performance.

\subsection{Audio Length Distribution Analysis}
\label{subsec:audio_length_distribution}

Table~\ref{tab:audio_length_distribution} reports the distribution of audio sample lengths in the dataset. Notably, 216 samples (approximately 17\%) have a duration longer than 15 seconds.

\begin{table*}[]
\centering
\begin{tabular}{l|l|l}
\toprule
\textbf{Hyperparam}   & \textbf{ContextPool-RoBERTa (N=1)}        & \textbf{ContextPool-RoBERTa (N=5)}        \\ \hline
Backbone              & RoBERTa large           & RoBERTa large           \\
Model Type            & ContextPoolingTextModel & ContextPoolingTextModel \\
Hidden Layers         & {[}100, 50{]}           & {[}100, 50{]}           \\
Dropout               & 0.1                     & 0.1                     \\
Context Window        & 1                       & 5                       \\
Optimizer             & AdamW                   & AdamW                   \\
Learning Rates        & 3.4e-5                  & 1.5e-5                  \\
Weight Decay          & 8.05e-5                 & 3.9e-7                  \\
Learning Rate Decay   & Linear                  & Linear                  \\
Warmup Steps          & 258                     & 258                     \\
Batch Size            & 8                       & 4                       \\
Gradient Accumulation & 3                       & 3                       \\
Max Steps             & 860                     & 860                     \\
Precision             & bf16-mixed              & bf16-mixed              \\
Seed                  & 20                      & 20                      \\
\bottomrule
\end{tabular}
\caption{Hyperparameters of our best text-only models.}
\label{tab:hyperparams_text}
\end{table*}

\begin{table*}[]
\centering
\begin{tabular}{l|l}
\toprule
\textbf{Hyperparam}       & \textbf{HuBERT-Base fine-tuned}          \\ \hline
Backbone                  & HuBERT base                  \\
Model Type                & BaseModel                    \\
Hidden Layers             & {[}50{]}                     \\
Layer to Finetune         & 3                            \\
Dropout                   & 0.1                          \\
Context Window            & 0                            \\
Optimizer                 & AdamW                        \\
Learning Rates            & 2e-4 (backbone), 2e-4 (head) \\
Weight Decay              & 0.01                         \\
Learning Rate Decay       & Linear                       \\
Warmup Steps              & 258                          \\
Batch Size                & 4                            \\
Gradient Accumulation     & 3                            \\
Max Steps                 & 860                          \\
Precision                 & bf16-mixed                   \\
Seed                      & 20                           \\        
\bottomrule
\end{tabular}
\caption{Hyperparameters of our audio model.}
\label{tab:hyperparams_audio}
\end{table*}

\begin{table*}[]
\begin{minipage}{0.45\linewidth}
\centering
\begin{tabular}{l|l}
\toprule
\textbf{Model}                   & \textbf{Weight} \\ \hline
ContextPool-RoBERTa (N=1)        & 0.4256          \\
ContextPool-RoBERTa (N=1)        & 0.3723          \\
ContextPool-RoBERTa (N=5)        & 0.2021          \\
\bottomrule
\end{tabular}
\caption{Weights for ensemble predictions of text-only models.}
\label{tab:hyperparameters_ensemble_text}
\end{minipage}
\hspace{0.5 cm}
\begin{minipage}{0.45\linewidth}
\centering
\begin{tabular}{l|l}
\toprule
\textbf{Model} & \textbf{Weight} \\ \hline
ContextPool-RoBERTa (N=1)           & 0.8128          \\
HuBERT-Base fine-tuned          & 0.1872          \\
\bottomrule
\end{tabular}
\caption{Weights for ensemble predictions of best text and audio models.}
\label{tab:hyperparameters_ensemble_text_audio}
\end{minipage}
\end{table*}

\begin{table*}[]
\centering
\begin{tabular}{ll}
\toprule
\textbf{Length Interval (s)}      & \textbf{Number of Samples} \\ \hline
0--1                              & 18                         \\
1--3                              & 201                        \\
3--5                              & 231                        \\
5--10                             & 399                        \\
10--15                            & 213                        \\
15+                               & 216                        \\
\bottomrule
\end{tabular}
\caption{Distribution of audio samples by length intervals (in seconds).}
\label{tab:audio_length_distribution}
\end{table*}

\begin{table*}[]
\centering
\begin{tabular}{llll}
\toprule
\textbf{Class} & \textbf{Avg. Length (s)} & \textbf{Std (s)}    & \textbf{Max Length (s)} \\ \hline
\textbf{AE}    & 9.11                     & 8.88                & 123.79                  \\
\textbf{AA}    & 10.80                    & 12.10               & 137.27                  \\
\textbf{AH}    & 9.14                     & 10.49               & 83.98                   \\
\textbf{FC}    & 11.28                    & 7.16                & 39.75                   \\
\textbf{SS}    & 10.64                    & 7.45                & 46.00                   \\
\textbf{S}     & 8.97                     & 10.69               & 39.59                   \\ 
\bottomrule
\end{tabular}
\caption{Length distribution statistics (average, standard deviation, and maximum) of audio samples for each class.}
\label{tab:length_distribution_classes}
\end{table*}

\begin{table*}[]
\centering
\begin{tabular}{ll}
\toprule
\textbf{Context Window} & \textbf{Average Duration (s)} \\ \hline
1                       & 5.74                          \\
2                       & 11.29                         \\
3                       & 16.63                         \\
4                       & 21.76                         \\
5                       & 26.69                         \\
6                       & 31.44                         \\
\bottomrule
\end{tabular}
\caption{Average duration of the context as a function of the context window size.}
\label{tab:avg_duration_context}
\end{table*}

\end{document}